\documentclass[letterpaper]{article} % DO NOT CHANGE THIS
\usepackage{aaai2026}  % DO NOT CHANGE THIS
\usepackage{times}  % DO NOT CHANGE THIS
\usepackage{helvet}  % DO NOT CHANGE THIS
\usepackage{courier}  % DO NOT CHANGE THIS
\usepackage[hyphens]{url}  % DO NOT CHANGE THIS
\usepackage{graphicx} % DO NOT CHANGE THIS
\usepackage{natbib}  % DO NOT CHANGE THIS AND DO NOT ADD ANY OPTIONS TO IT
\usepackage{caption} % DO NOT CHANGE THIS AND DO NOT ADD ANY OPTIONS TO IT
\usepackage{algorithm}
\usepackage{algorithmic}
\usepackage{amsmath}
\usepackage{amssymb}
\usepackage{booktabs}
\usepackage{pifont}
\usepackage{makecell}
\usepackage{multirow}

\usepackage{newfloat}
\usepackage{listings}
\DeclareCaptionStyle{ruled}{labelfont=normalfont,labelsep=colon,strut=off} % DO NOT CHANGE THIS
\floatstyle{ruled}
\newfloat{listing}{tb}{lst}{}
\floatname{listing}{Listing}
\title{BuildingWorld: A Structured 3D Building Dataset for Urban Foundation Models}
\author{
    Shangfeng Huang\textsuperscript{\rm 1},
    Ruisheng Wang\textsuperscript{\rm 2}\equalcontrib,
    Xin Wang\textsuperscript{\rm 1}\equalcontrib,
}
\affiliations{
    \textsuperscript{\rm 1}University of Calgary\\
    \textsuperscript{\rm 2}Shenzhen University\\
    {shangfeng.huang, xcwang}@ucalgary.ca, 
    ruiswang@szu.edu.cn
}

\usepackage{bibentry}
\makeatletter
\newif\ifreproStandalone
\reproStandalonefalse
\makeatother

\begin{document}

\maketitle
%\section*{Reproducibility Checklist}

\begin{abstract}
As digital twins become central to the transformation of modern cities, accurate and structured 3D building models emerge as a key enabler of high-fidelity, updatable urban representations. These models underpin diverse applications including energy modeling, urban planning, autonomous navigation, and real-time reasoning. 
Despite recent advances in 3D urban modeling, most learning-based models are trained on building datasets with limited architectural diversity, which significantly undermines their generalizability across heterogeneous urban environments. 
To address this limitation, we present \textbf{BuildingWorld}, a comprehensive and structured 3D building dataset designed to bridge the gap in stylistic diversity. 
It encompasses buildings from geographically and architecturally diverse regions—including North America, Europe, Asia,  Africa, and Oceania—offering a globally representative dataset for urban-scale foundation  modeling and analysis. 
Specifically, BuildingWorld provides about \textbf{five million LOD2 building models} collected from diverse sources, accompanied by \textbf{real and simulated airborne LiDAR point clouds}. This enables comprehensive research on 3D building reconstruction, detection and segmentation. 
Cyber City, a virtual city model, is introduced to enable the generation of unlimited training data with customized and structurally diverse point cloud distributions.
Furthermore, we provide standardized evaluation metrics tailored for building reconstruction, aiming to facilitate the training, evaluation, and comparison of large-scale vision models and foundation models in structured 3D urban environments.
\end{abstract}

    % \link{Code}{https://aaai.org/example/code}
    % \link{Datasets}{https://shangfenghuang.github.io/BuildingWorld}
    % \link{Extended version}{https://aaai.org/example/extended-version}

\section{Introduction}
%------------------------------------ Explain the word "LOD2" ------------------------------------------------------
\label{sec:intro}
\begin{figure}[h]
    \centering
    \includegraphics[width=1\linewidth]{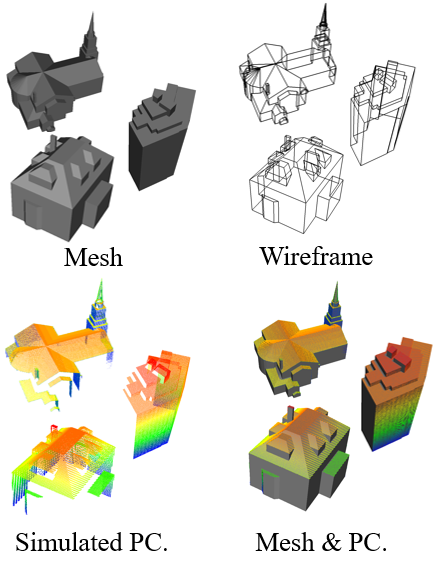}
    \caption{A glimpse of the BuildingWorld dataset.}
    \label{fig:1}
\end{figure}
As urban digitization accelerates, the development of high-fidelity and continuously updatable 3D building models has become a cornerstone in enabling digital twin cities \cite{deng2021systematic}. In digital twins, building models serve not only as geometric representations but also as integrative platforms that fuse heterogeneous urban data—ranging from energy use and structural integrity to mobility patterns and environmental conditions. These models have already demonstrated significant utility in a range of critical applications, including energy simulation \cite{pan2023building}, urban planning
\cite{ernst2021virtual}, emergency response \cite{demir20163d}, and virtual reality environments \cite{zhang2021urbanvr}. 
In the past decade, the community has made remarkable progress in building reconstruction from various modalities, such as aerial imagery\cite{tack20123d}, LiDAR point clouds \cite{bauchet2020kinetic, nan2017polyfit, huang2022city3d, li2022ransac, wang2020robust}, multi-view stereo (MVS) \cite{yu2021automatic, luo2024large}. 
Recently, building reconstruction datasets with annotations and benchmarks \cite{peralta2020next,selvaraju2021buildingnet,wang2023building3d,Tokyo} have significantly accelerated progress in this field, providing standardized resources for model training and evaluation.  
While these datasets have contributed significantly to progress in 3D building reconstruction, they often remain limited in architectural diversity and geographic scope, which constrains the generalizability of learned models \cite{liu2024point2building, huang2024pbwr, hao2025edge}.
% The accuracy and semantic richness of building models directly influence the fidelity and utility of digital twin systems. 

\begin{table*}[t]
  \centering
  % \small  
  \setlength{\tabcolsep}{1mm} 
  \renewcommand{\arraystretch}{1.3}
  \begin{tabular}{cccccccc}
    \toprule
    Dataset & Scene & Sensors & \#Models & Diversity & PC & Mesh & WF \\
    \midrule
    RoofN3D \shortcite{wichmann2018roofn3d} & City & ALS & -- & New York (1,009 $km^2$) & \ding{51} & \ding{51} & \ding{55} \\
    DublinCity \cite{zolanvari2019dublincity} & Urban & ALS & -- & Dublin (5.6 $km^2$) & \ding{51} & \ding{55} & \ding{55} \\
    Houses3K \cite{peralta2020next} & Object & Handcraft & 3\,K & 600 unique buildings & \ding{55} & \ding{51} & \ding{55} \\
    3DBAG \cite{leon2021testing} & Country & ALS & 10\,M & Netherlands & \ding{51} & \ding{51} & \ding{55} \\
    SUM \cite{gao2021sum} & Urban & Aerial Photogram. & -- & Helsinki (4 $km^2$) & \ding{55} & \ding{51} & \ding{55} \\
    BuildingNet \cite{selvaraju2021buildingnet} & Object & Handcraft & 2\,K & -- & \ding{55} & \ding{51} & \ding{55} \\
    City3D \cite{huang2022city3d} & Urban & ALS & 20\,K & 3 cities & \ding{51} & \ding{51} & \ding{55} \\
    UrbanScene3D \cite{lin2022capturing} & Urban & MVS & 1.4\,K & \makecell{10 synthetic /\\ 6 real scenes (55 $km^2$)} & \ding{55} & \ding{51} & \ding{55} \\
    STPLS3D \cite{chen2022stpls3d} & -- & UAV Photogram. & -- & \makecell{63 synthetic /\\ 4 real scenes (16 $km^2$)} & \ding{51} & \ding{51} & \ding{55} \\
    SensatUrban \cite{hu2022sensaturban} & Urban & UAV Photogram. & -- & UK cities (7.6 $km^2$) & \ding{51} & \ding{55} & \ding{55} \\
    Building3D \shortcite{wang2023building3d} & Country & ALS & 760\,K & Estonia & \ding{51} & \ding{51} & \ding{51} \\
    \midrule
    BuildingWorld & Urban (Global) & \makecell{ALS\\Simulated ALS} & 5\,M & Five continents& \ding{51} & \ding{51} & \ding{51} \\
    \bottomrule
  \end{tabular}
  \caption{Comparison with representative 3D building datasets. \textit{K}: thousand, \textit{M}: million, \textit{PC}: point clouds, \textit{WF}: wireframe, \textit{ALS}: airborne laser scanning, MVS: multi-view stereo, \textit{Photogram.}: photogrammetry.}
  \label{tab:dataset}
\end{table*}

To this end, we propose \textbf{BuildingWorld}, a structured 3D building dataset for urban foundation models, designed to enhance the generalization capabilities of deep learning-based models. 
The emergence of large language models, such as ChatGPT \cite{achiam2023gpt}, DeepSeek \cite{liu2024deepseek, guo2025deepseek}, and LLaMA \cite{touvron2023llama, touvron2023llama2}, has demonstrated the critical role of large-scale, high-quality datasets in enabling strong reasoning and generalization capabilities. 
Similarly, large vision models, including SAM \cite{kirillov2023segment} and Dinov2 \cite{oquab2023dinov2}, have achieved impressive performance by mining knowledge from massive image datasets.
This paradigm highlights the need for similarly large-scale and diverse datasets in other domains, such as 3D urban modeling, to unlock comparable levels of generalization and reasoning.
Therefore, the BuildingWorld dataset is constructed from about five million LOD2 building models, along with both real and simulated airborne LiDAR point clouds. These building models are collected from diverse sources and are distributed across geographically and architecturally varied regions, including North America, Europe, Asia, Africa, and Oceania. This wide coverage theoretically encompasses most major building styles found around the world.
The BuildingWorld dataset aims to overcome the limitations imposed by biased data distributions and limited architectural diversity, which often hinder the generalization capabilities of current models.

Compared with text and image data, which are abundant and easily accessible on the web, the collection of 3D data, especially LiDAR point clouds, is labor-intensive, costly, and rarely available to the public. 
Recently, several advanced large point cloud–language models\cite{ i2024gpt4point, xu2024pointllm} try to generate large amounts of synthetic text–point cloud pairs to support the training and evaluation of large models. In parallel, some methods \cite{huang2025arcpro, otsuka2025pre} simulate point clouds to enhance model performance. 
Motivated by them, BuildingWorld generates simulated aerial LiDAR point clouds from LOD2 building models to approximate real-world point cloud acquisition. This allows model trained on BuildingWorld to better generalize to real-world LiDAR data. 
Specifically, the LiDAR simulator tool in Helios++ \cite{winiwarter2022virtual} is used to simulate the acquisition of aerial point clouds, accounting for real-world factors such as occlusion, laser incidence angle, flight trajectory, and other sources of point cloud incompleteness. 
In addition, complete building point clouds are made available to support specific tasks, such as point cloud completion.
Furthermore, to generate more realistic simulated aerial point clouds, infrastructure elements, terrain models, and vegetation such as trees are incorporated into the virtual environments in some cities.

Beyond standard benchmarks, we introduce supplementary evaluation metrics designed to more accurately assess reconstruction accuracy, completeness, and robustness under complex building conditions. 
A benchmark is conducted using five representative deep learning methods, validating the potential and utility of the dataset in the building reconstruction task.
Furthermore, we analyze the limitations and challenge present in the dataset, and outline a range of potential downstream applications in digital twins, urban simulation, and large-scale 3D reconstruction.

Our contributions are summarized as following:
\begin{itemize}
    \item We propose BuildingWorld, the first and largest structured 3D building dataset for urban foundation models, which contains about five million LOD2 building models from geographically and architecturally diverse regions spanning five continents, including North America, Europe, Asia, Africa and Oceania.
         
    \item We establish a comprehensive benchmark with standard and supplementary evaluation metrics, designed to assess reconstruction accuracy, completeness, and robustness.
    \item We create an 3D urban scene generator called Cyber City to generate 3D urban models (buildings, trees etc.) and 3D point clouds to enrich 3D dataset for large 3D foundation models.
    \item We simulate aerial point clouds using the Helios++ LiDAR simulator, incorporating realistic factors such as occlusion, laser incidence angle, and flight trajectory.  Real aerial point clouds are also provided as part of the dataset to support evaluation and validate the effectiveness of models trained on simulated data.     
\end{itemize}

\begin{figure*}[t]
    \centering
    \includegraphics[width=\linewidth]{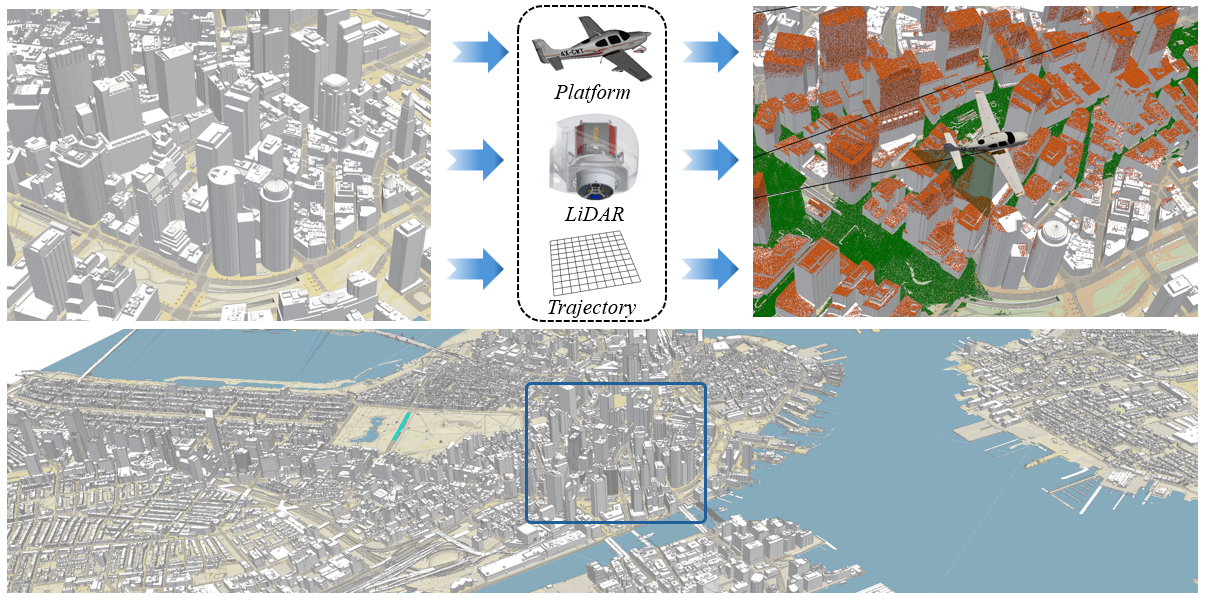}
    \caption{Illustration of the construction process of BuildingWorld dataset. A glimpse of the LOD2 digital city model of Boston is shown. The zoomed-in downtown area illustrates simulated aerial LiDAR point clouds, generated using a predefined airborne platform, LiDAR sensor, and flight trajectory.}
    \label{fig:BuildingWorld}
\end{figure*}

\section{Related Work}

\subsection{Related Datasets}

In recent years, the increasing availability of annotated building datasets has significantly promoted the development of deep learning-based 3D building reconstruction methods, as summarized in \ref{tab:dataset}. 
RoofN3D \cite{wichmann2018roofn3d}, 3DBAG \cite{leon2021testing}, STPLS3D \cite{chen2022stpls3d} and City3D \cite{huang2022city3d} provide large-scale aerial LiDAR point clouds and building mesh models generated through automatic reconstruction methods. However, these building models often exhibit geometric errors in local details, which makes them unsuitable as ground truth for supervised learning methods. 
DublinCity \cite{zolanvari2019dublincity} and SensatUrban \cite{hu2022sensaturban} are two recently popular urban point cloud datasets for semantic segmentation, but they lack high-quality building mesh models. Houses3K \cite{peralta2020next}, UrbanScenes3D \cite{lin2022capturing} and BuildingNet \cite{selvaraju2021buildingnet} provide high-quality, manually crafted building models with limited architectural diversity, whereas SUM \cite{gao2021sum} offers low-quality urban meshes primarily used for mesh semantic segmentation. 
Due to low mesh quality or the absence of corresponding point clouds, the aforementioned datasets are rarely used for training supervised methods for building reconstruction from point clouds.
Building3D \cite{wang2023building3d} is the first dataset specifically designed for LOD2 building reconstruction from aerial LiDAR point clouds. Notably, the provided wireframe models are manually created and well suited for training deep learning-based methods. 
However, these datasets are geographically limited, which constrains the generalization and robustness of models trained on them.
To address this limitation, we propose BuildingWorld, the first building reconstruction dataset with architectural diversity spanning five continents, including North America, Europe, Asia, Africa and Oceania.

\subsection{Related Methods}

Traditional 3D building reconstruction methods can be divided into model-driven and data-driven approaches based on their reconstruction strategies.
Model-driven methods \cite{zhang2021optimal, song2020curved, li2022ransac, zang2024compound} typically rely on the assumption of structured building rules, where building models can be approximated by fitting simple parametric shapes, such as hip and flat roofs.  
Data-driven methods \cite{zhou20102, zhou20112, chen2017topologically, nan2017polyfit, yang2022connectivity} focus on extracting building primitives from point clouds and reconstructing the topological relationships between these primitives to obtain watertight models.
However, these traditional methods tend to be highly sensitive to parameter selection and often struggle to generalize to diverse building shapes and structural variations.
Recently, several deep learning-based methods \cite{li2022point2roof, yang2024method, hao2025edge, huang2024pbwr} have been proposed to reconstruct building models from point clouds. 
Specifically, heuristic methods \cite{li2022point2roof, yang2024self, jiang2023extracting, hao2025edge, hao2024leaf, wang2023building3d} treat building model reconstruction as a wireframe reconstruction task consisting of corners and edges. 
% These methods first detect corner points from the input point clouds, and then regress positive edges from a set of candidate edges generated by connecting each pair of corner points.  
% However, candidate edges are highly imbalanced, with many negative samples and few positive ones. 
However, heuristic reconstruction strategies inherently suffer from error accumulation across different stages. 
Therefore,  some methods \cite{luo2022learning, yang2024method, huang2024pbwr} propose to regress building edges directly from point clouds, without relying on corner detection. However, these methods still rely on post-processing to generate the final wireframe models. 
Recently, some new building  reconstruction paradigms are also proposed. Point2Building  \cite{liu2024point2building} adopts a generative strategy to reconstruct building models. EdgeDiff \cite{liu2025edgediff} leverages diffusion models to recover building structures from a noise field, while BWFormer \cite{liu2025bwformer} employs a powerful 2D corner detector to reduce the impact of corner detection errors on the subsequent edge regression module.

%%%%%%%%%% BuildingWorld Dataset
\begin{figure}[!t]
    \centering
    \includegraphics[width=1\linewidth]{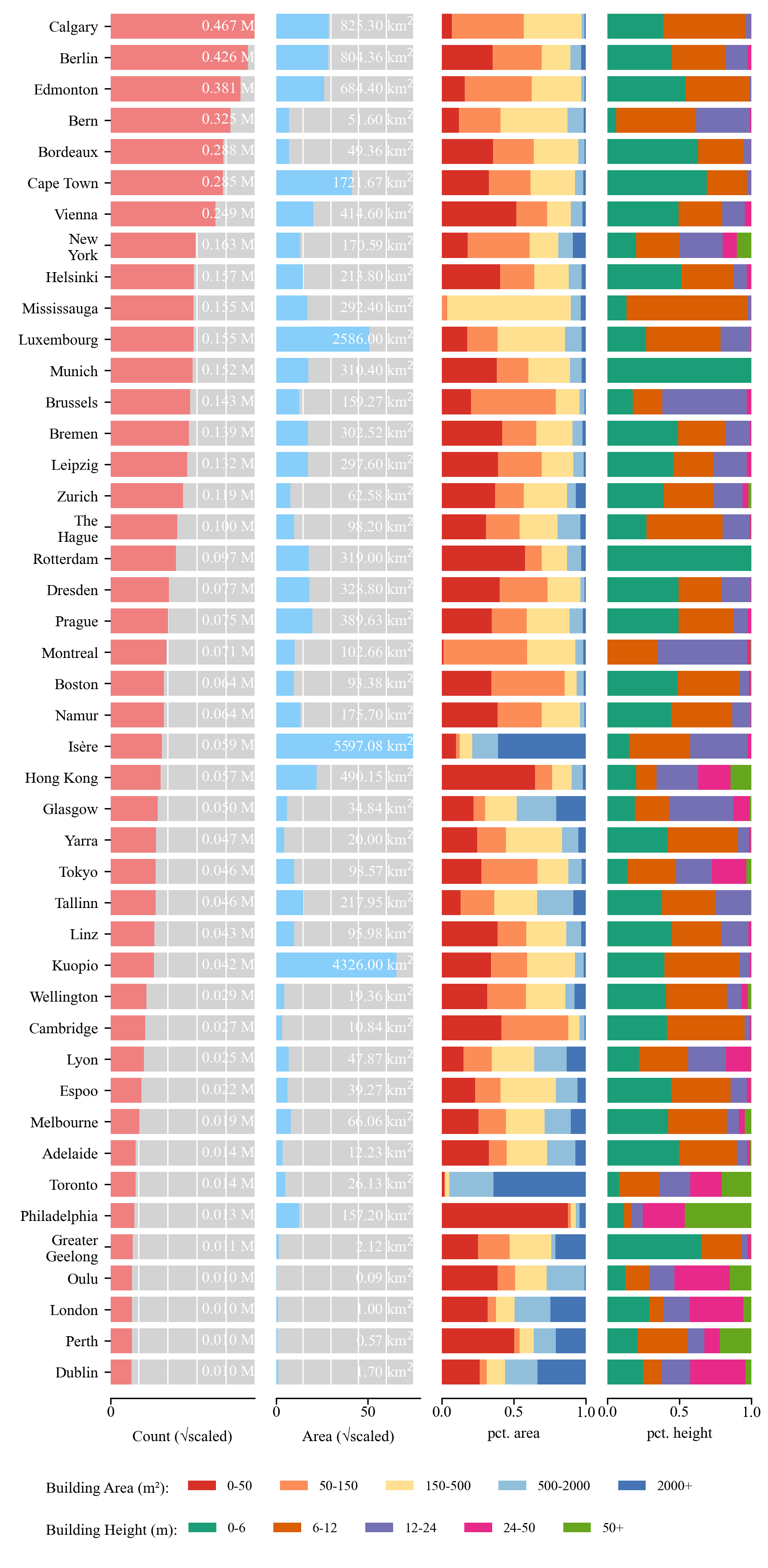}
    \caption{Statistical overview of the BuildingWorld dataset. Area bars indicate scene sizes, while percentage area and height metrics highlight the diversity of building structures.}
    \label{fig:data_distribution}
\end{figure}

\section{BuildingWorld Dataset}
\subsection{Overview}
The BuildingWorld dataset is constructed by collecting all publicly available LoD2 building models from around the world, with the goal of capturing diverse global architectural styles and mitigating limitations in model generalization caused by insufficient or imbalanced data distribution.
Aerial point clouds are subsequently simulated using the Helios++ simulator \cite{ernst2021virtual}, as shown in Figure \ref{fig:BuildingWorld}. The Helios++ simulator enables the generation of realistic aerial point clouds by modeling phenomena such as laser incidence angle effects, occlusions from buildings and trees, and other environmental interferences present in real-world data. 
To further diversify the data distribution, we construct a synthetic urban scene named \textit{Cyber City}, which integrates buildings of diverse styles and cultural origins from around the world, thereby breaking geographical and temporal constraints. Theoretically, it is possible to generate an infinite variety of Cyber City instances by randomly assembling building models with diverse styles and cultural backgrounds in different spatial configurations.

\subsection{Building Models}
To the best of our efforts, BuildingWorld dataset are constructed by collecting LoD2 building models from 44 cities, covering an area of  \textbf{21,718} km$^2$, resulting in a total of approximately 5 million buildings. These models span a wide variety of types and sources: some are derived from architectural blueprints provided by local governmental land management agencies, others are manually created, some are generated by integrating multiple data sources with predefined roof geometries, as shown in Figure \ref{fig:data_distribution}. To the best of our knowledge, this is the first publicly available dataset that provides such a comprehensive and globally diverse collection of LoD2 building models.  As illustrated in Figure \ref{fig:1}, the LoD2 building models exhibit greater geometric detail, including features such as chimneys and sharply pitched roofs. The roof geometry of LoD2 models serves as an important metric for evaluating the fidelity of individual building models and the overall quality of the dataset. 
% Analysis data %
In the North American subset of the dataset, urban areas are predominantly characterized by small-scale residential zones covering 50–150 km² and medium-scale regions, including both residential and public buildings, ranging from 150–500 km². The building height distribution is largely concentrated in the 0–6 meter range (single-story structures) and the 6–12 meter range (two- to three-story buildings), with high-rise structures appearing relatively infrequently. 
Such a distribution is consistent with the urban planning and demographic realities of these countries, where ample livable land and moderate population pressures have led to a prevalence of low-density, low-rise constructions. % this sentence can be deleted
In contrast, the Hong Kong subset exhibits a significantly higher proportion of high-rise buildings, predominantly in the 12–50 meter height range.
Figure \ref{fig:data_distribution} demonstrates the diversity of building models within the dataset through the distribution of building footprint area and the percentage breakdown of building sizes and heights.

\begin{figure*}
    \centering
    \includegraphics[width=1\linewidth]{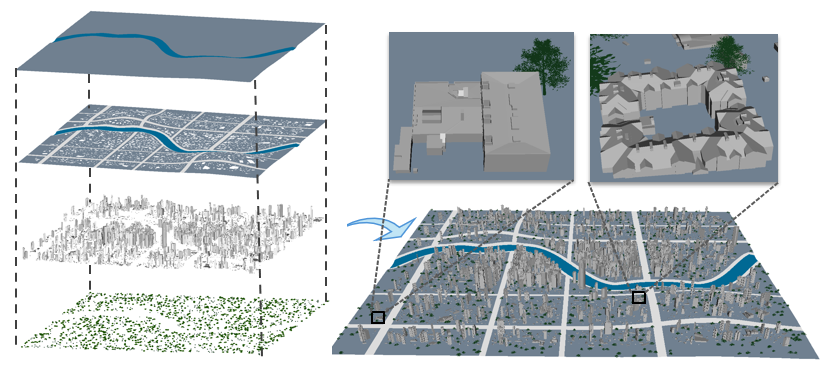}
    \caption{Illustration of Cyber City, which consists of four main components: terrain, road and building footprints, buildings, and vegetation.}
    \label{fig:cybercity}
\end{figure*}

\subsection{Simulated ALS}
Some cities agencies provide not only structurally detailed LoD2 building models but also corresponding real aerial point clouds. However, our analysis reveals that the geometric consistency between the models and the real point clouds is insufficient to support their direct use in supervised learning. This misalignment may arise from various factors, such as positional inaccuracies introduced during the modeling process—especially when models are created based on architectural blueprints—or from subtle yet impactful discrepancies in roof geometry, where the government-provided models exhibit deviations from the actual roof structures observed in point clouds.
Nonetheless, real aerial point clouds remain useful in reinforcement learning stage, especially when an effective reward-based ranking strategy is employed to guide the training of large 3D vision modules. Therefore, the processed real point clouds are also collected and  made available in BuildingWorld.

To construct a high-quality dataset for 3D building point cloud reconstruction, the Helios++ simulator \cite{ernst2021virtual} is employed to generate realistic aerial point clouds, as shown in Figure \ref{fig:BuildingWorld}.  
The simulator is composed of three main components—vehicle platform, LiDAR sensor, and flight trajectory—which jointly determine key properties of the generated point clouds, such as density, coverage, and structural completeness. Specifically, the vehicle platform primarily controls the flight speed, which directly influences point cloud density and the completeness of fine structural details. Lower flight speeds typically yield denser point clouds but increase the computational burden for model training. Conversely, higher speeds often result in a substantial loss of rooftop details—such as chimneys—due to sparser sampling. In practice, flight speeds are chosen based on the application scenario and task requirements. In BuildingWorld, point clouds for each scene are simulated using a randomly selected speed within the range of 185–463 km/h (100–250 knots), rather than a fixed value, to enhance the diversity of point cloud distributions. 
The RIEGL VQ-1560 II-S, one of the most advanced and widely used airborne LiDAR sensors, is selected as the simulated sensor in our framework. During simulation, the laser pulse repetition rates are set to 1M and 2M Hz, with a scan angle of ±30°. The scan frequencies are defined as 200, 400, and 600 Hz. These parameters are selected in combination with varying flight speeds to generate diverse point cloud distributions. 
Pulse repetition rate and scan frequency directly determine the density of the generated point clouds. As these values increase, the resulting point clouds capture finer rooftop details and contain a greater number of points.
The flight trajectory plays a critical role not only in determining the overall density of the point cloud, but also in shaping the spatial distribution of occlusions—areas where parts of buildings are missing due to line-of-sight obstructions. In our experiments, the primary flight trajectory is aligned in the north-south direction, while the secondary trajectory follows an east-west orientation. During the simulation, the main flight path is oriented north–south, with an auxiliary trajectory in the east–west direction. To reduce occlusion-related point cloud loss, a lateral overlap rate between $40\% $ and $60\%$ is maintained. 
In addition, the altitude of the flight trajectories is set within a range of 600 to 1200 meters above sea level. Together, the flight trajectory and altitude control the distribution of point cloud coverage on building facades. At higher altitudes, the laser incidence angle relative to vertical surfaces becomes smaller, resulting in reduced interaction with facades and, consequently, lower completeness of facade point clouds compared to lower-altitude flights. Nevertheless, facade points reconstructed from aerial LiDAR data using LoD2 building models can significantly improve the estimation of building height.

BuildingWorld provides  simulated aerial point clouds generated via Helios++, as shown in Figure \ref{fig:1}.  In addition, complete point clouds also are sampled directly from the 3D building models, offering comprehensive geometric data for specific tasks.
For example, when training a large-scale foundation model for 3D building point clouds, aerial point clouds generated by Helios++ can be used as input, while the completed point clouds predicted by the model are supervised with ground-truth geometry sampled directly from the original building models.
Specifically, uniform sampling is employed to generate points from the building models, ensuring a consistent point density of 30 $points / m^2$. Furthermore, to simulate realistic sensing noise, random perturbations are introduced along both the planar surface and normal directions during sampling, enhancing the natural variability of the resulting point clouds.

\subsection{Cyber City}

Across continents, architectural styles manifest distinct regional characteristics. European cities emphasize contextual modernism that integrates with historical environments. African urban landscapes reflect a coexistence of colonial-era structures and modernist glass forms, balancing cultural heritage with pragmatic functionality. Asian cities are defined by high-density, technologically driven verticality. North American architecture favors minimalist functionalism and spatial openness, while Oceania prioritizes climate-adaptive, sustainable design. 
Unbound by geographical location or cultural history, Cyber City serves as a virtual city where building models representing diverse architectural styles from all continents are brought together to form a globally inclusive urban composition, as shown in Figure \ref{fig:cybercity}. Being procedurally generated, it can produce unlimited synthetic urban configurations with diverse architectural compositions, facilitating richer data distributions for model training.

Cyber City consists of four main layers: terrain, road and building footprints, vegetation, and 3D building.  In the terrain layer, we begin by randomly defining the spatial boundaries of the city, which serves as the basis for subsequent urban layout generation, as shown \ref{fig:cybercity} . It defines an area of 4 × 4 $km$. Given that rivers frequently play a central role in the formation of real-world cities, a primary river is randomly generated to run through the synthetic urban area. Furthermore, to enhance terrain diversity, random local protrusions are applied to simulate small hills across the ground surface. 
The footprint layer defines the spatial organization of roads and buildings within the city. Streets are primarily arranged in an orthogonal grid pattern, while building footprints are allocated to the remaining parcels, specifying the exact locations for building placement.
In Building layer, building area and height are used as heuristic indicators to differentiate functional types of buildings. As shown in Figure \ref{fig:cybercity}, the central region is designated as the Central Business District, where taller structures are placed to represent high-density commercial buildings, typically surrounded by lower-rise residential areas. In recreational and public activity zones, such as shopping malls and libraries, building models with relatively large footprint areas are placed to reflect their functional requirements. It is important to note that this configuration reflects conventional urban logic, but a freely designed virtual city imposes no such constraints—its structure can be arbitrarily defined to suit any desired purpose or experimental condition. 
In the tree layer, vegetation is distributed in a space-filling manner, with trees placed in unoccupied or residual areas of the scene to complement the urban layout. Additionally, the types and sizes of trees are randomly selected and generated to increase variability and naturalness in the scene. 
The tree layer is incorporated to simulate occlusions commonly encountered during point cloud acquisition, where trees partially obstruct building structures. Such occlusions are frequently observed in real-world urban environments.

\begin{table*}
    \centering
    \begin{tabular}{>{\centering}p{5cm}|c|>{\centering}p{1.3cm} >{\centering}p{1.3cm} >{\centering}p{1.3cm}|>{\centering}p{1.3cm}  >{\centering}p{1.3cm} c}
    \toprule
          \multirow{2}{*}{Method}&  Distance (m)&  \multicolumn{6}{c}{Accuracy}\\
  & ACO & CP& CR& $\text{CF}_1$& EP& ER&$\text{EF}_1$\\
   \midrule
    PBWR *  \cite{huang2024pbwr}& 0.27& 0.95& 0.67& 0.78& 0.84& 0.61&0.71\\

    \midrule
    PBWR  \cite{huang2024pbwr}& 0.22& 0.96& 0.68& 0.80& 0.91& 0.65&0.76 \\
    PBWR * \cite{huang2024pbwr}& 0.36& 0.85& 0.65& 0.74& 0.71& 0.60&0.65\\
    \bottomrule
    \end{tabular}
    \caption{Performance comparison of models on simulated and real-world point clouds. The first row reports performance on simulated data, while the second and third rows show results on real aerial point clouds. * indicates models trained on simulated point clouds; otherwise, models are trained on real point clouds.}
    \label{Tab:Tallinn}
\end{table*}

% Cyber City is procedurally generated, allowing for the creation of an unlimited number of synthetic urban configurations with diverse architectural compositions. Leveraging the Helios++ point cloud simulator, BuildingWorld can generate varied point cloud distributions by altering terrain shapes, building layouts, and vegetation placement. This enables models trained on the dataset to learn robust, spatially consistent representations across diverse urban forms.

%%%%%%%%%%  Benchmark
\section{Benchmark}

\subsection{Baseline Methods}
% Existing methods often exhibit poor robustness due to their heavy reliance on the distribution of the training dataset. 
% For instance, models trained on the Calgary city dataset frequently fail to generalize to urban scenes from other cities, where data distributions differ significantly. 
BuildingWorld is the first dataset designed to support the training of 3D foundation models for building reconstruction. 
However, due to the lack of publicly available large-scale 3D building reconstruction models, a direct evaluation of BuildingWorld’s effectiveness on such foundation models is currently not possible.
Given this limitation, we instead design a experiment to investigate how well models trained on BuildingWorld generalize to real-world aerial point clouds.

\subsection{Datasets and Experimental Setup}
Building3D \cite{wang2023building3d} dataset provides real aerial point clouds paired with manually labeled LoD2 building models of  Tallinn City, serving as ground truth for evaluation. In our experiments, we first train the baseline method using the simulated building data derived from Building3D, and subsequently evaluate its performance on the test set, which contains real aerial point clouds along with corresponding annotated building models.
This setup allows us to assess how well models trained solely on synthetic data transfer to real-world inputs. 
To establish a reference point and assess the generalization ability, we compare performance with the model trained directly on the Building3D dataset.
We follow the same evaluation metrics as defined in the Building3D dataset, as shown in Table \ref{Tab:Tallinn}. Specifically, ACO denotes the Average Corner Offset, measuring the mean positional deviation between predicted and ground-truth corner points, expressed in meters. CP and EP refer to the precision of predicted corners and edges, respectively, while CR and ER represent their corresponding recall values. $CF_1$ and $EF_1$ indicate the $F_1$-scores for corner and edge detection, respectively.

\subsection{Results and Discussion}
Table \ref{Tab:Tallinn} compares the performance of PBWR \cite{huang2024pbwr} trained on real data and simulated data, evaluated on both real and simulated point clouds. Experimental results indicate that the model trained on real data achieves performance comparable to that of the model (*) trained on simulated data when tested within their respective domains. 
% These results indicate that PBWR, originally designed for the Building3D dataset, generalizes effectively to simulated point clouds, highlighting its resilience to data distribution shifts within identical scenes.
It is noteworthy that the model (*) trained on simulated data also performs competitively on real-world point clouds, showing no substantial degradation in reconstruction accuracy.
Additionally, the real point cloud data used for training the original PBWR model includes both RGB and intensity information, which contribute to improved reconstruction performance. 
In contrast, the model (*) trained on simulated data relies solely on the spatial structure of point clouds for building reconstruction, yet still demonstrates strong applicability to real-world data. 
These findings indicate that models trained on BuildingWorld, using globally collected building models and simulated point clouds, are fully capable of supporting real-world applications. 
They further suggest that large-scale 3D building reconstruction models trained on BuildingWorld have strong potential to generalize effectively and accurately reconstruct buildings across diverse geographic regions worldwide.

%%%%%%%%%% Additional Task
\section{Challenges and Tasks}
% Although BuildingWorld is specifically designed as a large-scale, cross-regional dataset for training foundation models in LoD2 building reconstruction, it is also well-suited for a range of other 3D vision tasks, including the following:

\subsection{LoD3 Building Reconstruction}
Beyond the comprehensive LoD2 coverage, BuildingWorld also incorporates detailed LoD3 building models for selected cities, made available through contributions from government bodies or city planning authorities, such as Glasgow and Hong Kong. 
Inspired by the design concept of Cyber City, scattered LoD3 building models can be aggregated into a unified LoD3 Cyber City. Furthermore, by leveraging the Helios++ simulator, we can separately simulate façade point clouds using vehicle-mounted LiDAR sensors and rooftop point clouds using airborne LiDAR. The integration of these complementary point cloud sources enables new opportunities for exploring LoD3 building reconstruction.

\subsection{Semantic and Instance Segmentation}
% Due to the high cost of manual annotation, large-scale, city-level point cloud datasets for semantic segmentation remain relatively scarce. 
% However, with the rapid advancement of technologies such as autonomous driving and embodied intelligence, the development and progress of point cloud semantic and instance segmentation have become increasingly important. 
In BuildingWorld, all point clouds are annotated with unique semantic and instance-level labels, covering façades, roofs, buildings, vegetation, bridges, and terrain. Following the Cyber City design philosophy, additional elements such as vehicles and infrastructure models can be incorporated to further enrich the urban scene. Although replicating material-dependent artifacts in simulated point clouds remains challenging, BuildingWorld serves as a practical and effective resource for advancing city-scale semantic and instance segmentation tasks.

% \subsection{Object Detection}
% Its rich instance-level annotations and structural diversity open up new opportunities for exploring detection tasks in large-scale 3D urban environments. For example, one can investigate the detection of buildings, trees, and other city elements under varying occlusion, density, and spatial arrangements. 

% Beyond computer vision tasks, the city-scale building models provided by BuildingWorld can also support applications such as rooftop solar illumination analysis, UAV path planning, and other urban simulation scenarios.

%%%%%%%%% Conclusion
\section{Conclusion}
In this work, we introduce BuildingWorld, a large-scale, structured dataset tailored for 3D building understanding across diverse geographic and architectural contexts. By aggregating about five million LoD2 building models from 44 cities worldwide, and providing both real and simulated aerial LiDAR point clouds, BuildingWorld offers a unique resource for training and evaluating urban-scale 3D vision systems. The integration of the Cyber City framework further enables the generation of procedurally diverse synthetic scenes. 
We demonstrate that models trained exclusively on simulated data from BuildingWorld achieve competitive performance on real-world datasets, validating BuildingWorld’s effectiveness in bridging the synthetic-to-real gap.

\bibliography{aaai2026}

\end{document}